\documentclass[conference]{IEEEtran}
\IEEEoverridecommandlockouts
\usepackage{cite}
\usepackage{amsmath,amssymb,amsfonts}
\usepackage{algorithmic}
\usepackage{graphicx}
\usepackage{textcomp}
\usepackage{xcolor}
\def\BibTeX{{\rm B\kern-.05em{\sc i\kern-.025em b}\kern-.08em
    T\kern-.1667em\lower.7ex\hbox{E}\kern-.125emX}}

\usepackage{booktabs}
\usepackage{listings}
\usepackage{scalerel}
\usepackage[binary-units]{siunitx}
\usepackage{tablefootnote}
\usepackage[textsize=tiny]{todonotes}
\usepackage{tikz, pgfplots}
  \usetikzlibrary{svg.path}

\makeatletter
\let\NAT@parse\undefined
\makeatother
\usepackage{hyperref}

\newcommand\httpsurl[1]{%
  \href{https://#1}{\nolinkurl{#1}}%
}

\definecolor{orcidlogocol}{HTML}{A6CE39}
\tikzset{
  orcidlogo/.pic={
    \fill[orcidlogocol] svg{M256,128c0,70.7-57.3,128-128,128C57.3,256,0,198.7,0,128C0,57.3,57.3,0,128,0C198.7,0,256,57.3,256,128z};
    \fill[white] svg{M86.3,186.2H70.9V79.1h15.4v48.4V186.2z}
                 svg{M108.9,79.1h41.6c39.6,0,57,28.3,57,53.6c0,27.5-21.5,53.6-56.8,53.6h-41.8V79.1z M124.3,172.4h24.5c34.9,0,42.9-26.5,42.9-39.7c0-21.5-13.7-39.7-43.7-39.7h-23.7V172.4z}
                 svg{M88.7,56.8c0,5.5-4.5,10.1-10.1,10.1c-5.6,0-10.1-4.6-10.1-10.1c0-5.6,4.5-10.1,10.1-10.1C84.2,46.7,88.7,51.3,88.7,56.8z};
  }
}
\newcommand\orcidicon[1]{\href{https://orcid.org/#1}{\mbox{\scalerel*{
\begin{tikzpicture}[yscale=-1,transform shape]
\pic{orcidlogo};
\end{tikzpicture}
}{|}}}}

\makeatletter
\newcommand{\linebreakand}{%
  \end{@IEEEauthorhalign}
  \hfill\mbox{}\par
  \mbox{}\hfill\begin{@IEEEauthorhalign}
}
\makeatother

\newcommand\copyrighttext{%
  \footnotesize \textcopyright 2023 IEEE. Personal use of this material is permitted. Permission from IEEE must be obtained for all other uses, in any current or future media, including reprinting/republishing this material for advertising or promotional
  purposes, creating new collective works, for resale or redistribution to servers or lists, or reuse of any copyrighted component of this work in other works.
  DOI: \href{https://doi.org/10.1109/IV55152.2023.10186600}{10.1109/IV55152.2023.10186600}}
\newcommand\copyrightnotice{%
\begin{tikzpicture}[remember picture,overlay]
\node[anchor=south,yshift=10pt] at (current page.south) {\fbox{\parbox{\dimexpr\textwidth-\fboxsep-\fboxrule\relax}{\copyrighttext}}};
\end{tikzpicture}%
}
\begin{document}
\bstctlcite{IEEEexample:BSTcontrol}

\title{Combined Registration and Fusion of \\Evidential Occupancy Grid Maps for \\Live Digital Twins of Traffic\\
\thanks{\textsuperscript{\textdagger}These authors contributed equally.}
}

\author{\IEEEauthorblockN{Raphael van Kempen\textsuperscript{\orcidicon{0000-0001-5017-7494}\textdagger}}
\IEEEauthorblockA{\textit{Institute for Automotive Engineering} \\
\textit{RWTH Aachen University}\\
Aachen, Germany \\
raphael.vankempen@ika.rwth-aachen.de}
\and
\IEEEauthorblockN{Laurenz Adrian Heidrich\textsuperscript{\textdagger}}
\IEEEauthorblockA{\textit{Institute for Automotive Engineering} \\
\textit{RWTH Aachen University}\\
Aachen, Germany \\
laurenz.heidrich@rwth-aachen.de}
\and
\IEEEauthorblockN{Bastian Lampe\textsuperscript{\orcidicon{0000-0002-4414-6947}}}
\IEEEauthorblockA{\textit{Institute for Automotive Engineering} \\
\textit{RWTH Aachen University}\\
Aachen, Germany \\
bastian.lampe@ika.rwth-aachen.de}
\linebreakand
\IEEEauthorblockN{Timo Woopen\textsuperscript{\orcidicon{0000-0002-7177-181X}}}
\IEEEauthorblockA{\textit{Institute for Automotive Engineering} \\
\textit{RWTH Aachen University}\\
Aachen, Germany \\
timo.woopen@ika.rwth-aachen.de}
\and
\IEEEauthorblockN{Lutz Eckstein}
\IEEEauthorblockA{\textit{Institute for Automotive Engineering} \\
\textit{RWTH Aachen University}\\
Aachen, Germany \\
lutz.eckstein@ika.rwth-aachen.de}
}

\maketitle

\copyrightnotice

\begin{abstract}
Cooperation of automated vehicles (AVs) can improve safety, efficiency and comfort in traffic. Digital twins of Cooperative Intelligent Transport Systems (C-ITS) play an important role in monitoring, managing and improving traffic. Computing a live digital twin of traffic requires as input live perception data of preferably multiple connected entities such as automated vehicles (AVs). One such type of perception data are evidential occupancy grid maps (OGMs). The computation of a digital twin involves their spatiotemporal alignment and fusion. In this work, we focus on the spatial alignment, also known as registration, and fusion of evidential occupancy grid maps of multiple automated vehicles.
While there exists extensive research on the synchronization and fusion of object-based environment representations, the registration and fusion of OGMs originating from multiple connected vehicles has not been investigated much. We propose a methodology that involves training a deep neural network (DNN) to predict a fused evidential OGM from two OGMs computed by different AVs. The output includes an estimate of the first- and second-order uncertainty. We demonstrate that the DNN trained with synthetic data only outperforms a baseline approach based on coordinate transformation and combination rules also on real-world data. Experimental results on synthetic data show that our approach is able to compensate for spatial misalignments of up to 5 meters and 20 degrees.
\end{abstract}


\section{INTRODUCTION}

Non-connected AVs can only perceive as much of their environment as their on-board sensors allow them to. This is limited by the number of sensors available and visual restrictions, e.g. due to weather conditions or occlusions. Other traffic participants and infrastructural elements can obscure the line of sight and might thus lead to insufficient environment perception, e.g. on busy intersections or in construction sites. By sharing and fusing perception data between multiple AVs, both perception accuracy and reliability can be improved. This has been shown in the cloud-based collective environment model developed in the UNICAR\textit{agil} project~\cite{Lampe.2020} and is being extended to a live digital twin of traffic in the AUTOtech.\textit{agil} project~\cite{autotech}. This digital twin is created from data perceived by AVs as well as stationary and non-stationary infrastructure sensors.

OGMs describe the occupancy states of spatial cells in the environment of an AV. Evidential OGMs also quantify the first- and second-order uncertainty of the occupancy states. This uncertainty information shall be taken into account when fusing OGMs from multiple AVs. Combining spatial environment data requires the data to be temporally synchronized and spatially aligned. In this work, we assume the data to be temporally synchronized and focus on the spatial alignment and fusion. AVs can estimate their position and orientation by self-localization using e.g. GNSS. In real environments, this often comes with considerably high uncertainties in the estimated pose. Especially urban areas with high traffic volumes and many high buildings are a challenging task as GNSS signals can be obstructed. An approach that is able to align and fuse evidential OGMs with consideration of uncertain poses is thus an important feature for a live digital twin of traffic in a C-ITS.

Different approaches to fuse spatial occupancy information are proposed in recent works which, however, have some weaknesses. The authors of \cite{YUE.2016} and \cite{MENG.2022} fuse spatial information on the sensor level introducing high latencies. In \cite{LI.2018} and \cite{MARKIEWICZ.2018}, OGMs are spatially aligned using coordinate transformation without paying attention to pose uncertainties. Feature point matching across grid maps is used for spatial alignment in \cite{SUN.2018} and \cite{TANG.2021}, which both use manual feature extraction with decreasing efficiency in more complex OGMs.

This work presents a novel approach to combined OGM registration and fusion, which takes pose uncertainties into account. It utilizes a convolutional neural network (CNN) that performs the task of spatial alignment and fusion of occupancy information in one forward propagation step. We present a neural network architecture that takes two OGMs with corresponding, uncertain poses as inputs and predicts a fused OGM as output. Our contributions are 

\begin{itemize}
  \item a novel neural network architecture for combined and uncertainty-aware registration and fusion of evidential OGMs,
  \item a scalable method using physical-based rendering to generate realistic training data from simulation,
  \item a quantitative evaluation of the trained model in comparison to a rule-based baseline approach on synthetic data, and
  \item a qualitative evaluation of the model when presented with real-world data.
\end{itemize}

\section{BACKGROUND AND RELATED WORK}

This section introduces the OGMs being used in this work and briefly describes the mathematical concept of evidence theory as this is the basis for the occupancy information in the grid cells and the naive combination rule. Additionally, current approaches to OGM fusion are described.

\subsection{Occupancy Grid Maps}\label{ogm}

Object-based environment representations describe the existence probabilities and motion states of particular objects in the environment of an AV. On the other hand, an OGM~\cite{Elfes.1989} is a grid-based environment representation that does not rely on an object model but assigns occupancy information to discrete spatial cells in the environment of the AV. While probabilistic OGMs assign a scalar occupancy probability to each cell~\cite{Thrun.2005}, evidential OGMs use distinct belief masses, e.g. for the possible cell states "free" and "occupied". This is based on evidence theory as introduced by Dempster and Shafer~\cite{Dempster.1968,Shafer.1976} and allows to distinguish between aleatoric and epistemic uncertainty. This allows differentiating between those cells with an uncertain state because of missing data and those with an uncertain state because of conflicting data. The latter is especially important when data from different sources shall be fused, e.g. to create a live digital twin of traffic.

The \textbf{evidential OGMs} used in this work are based on the frame of discernment $\Theta = \{ F,O \}$, which comprises the possible and mutually exclusive cell states "free" and "occupied". Belief masses $m \in [0,1]$ are assigned to all possible subsets of $\Theta$, i.e. $2^\Theta = \{ \emptyset, \{ F \}, \{ O \}, \Theta \}$, while $\emptyset$ is not a possible cell state and $\Theta$ represents an unknown state.

Belief masses from different sources can be combined using \textbf{Dempster's Rule of Combination}. The formula given by~\cite{NUSS.2013} combines two belief masses $m_1$ and $m_2$ for the state of interest $X$:

\begin{equation}\label{combination}
\begin{split}
(m_1 \oplus m_2)(X) = \\ 
        \frac{\sum_{A,B \in 2^\Theta \mid A \cap B = X} m_1(A) \cdot m_2(B)}  {1-\sum_{A,B \in 2^\Theta \mid A \cap B = \emptyset} m_1(A) \cdot m_2(B)}
\end{split}
\end{equation}

It is possible to compute classical probability values from belief masses using \textbf{Pignistic Transform}:

\begin{equation}\label{eq:pignistic}
    p_O = m_O + 0.5 \cdot (1 - m_O - m_F)
\end{equation}

The belief masses in the cells of the OGM used in this work are predicted by a \textbf{deep inverse sensor model} as presented in \cite{VANKEMPEN.2021}. In that work, an end-to-end learning framework was developed to train a model to predict evidential OGMs from lidar measurements. It is based on the PointPillars architecture~\cite{Lang.2019b} and is trained using synthetic training data while still showing good generalization capability to real-world data.

\subsection{Fusion of Occupancy Grid Maps}

Occupancy information in OGMs perceived by different vehicles shall be combined into a single representation. This consists of two steps: spatial alignment and combination of occupancy information. There are several different existing approaches that can be grouped as follows:

\textbf{Early sensor fusion}: The approaches introduced in~\cite{YUE.2016} and~\cite{MENG.2022} both fuse the sensor data perceived by distinct agents directly and generate one OGM from the fused data afterwards. In \cite{YUE.2016}, two or more lidar point clouds are registered by an extension of the iterative closest point (ICP) algorithm, and in \cite{MENG.2022}, the point clouds are registered by coordinate transformation based on known poses. Drawbacks of this early fusion of sensor data are the processing efficiency due to high data volumes and the lack of a temporal synchronization between data sources.

\textbf{Coordinate transformation}: The approaches described in~\cite{LI.2018} and~\cite{MARKIEWICZ.2018} align two or more grid maps by using coordinate transformation only and then fuse the stored belief masses cellwise. The given poses are assumed to be perfectly accurate. The authors of~\cite{LI.2018} utilize a log-likelihood ratio for probability fusion, while~\cite{MARKIEWICZ.2018} compares several approaches such as Dempster's Rule of Combination, Bayesian Filtering, and Independent Opinion Pool. A drawback of these approaches is that real-world poses are never fully accurate, which is not considered with coordinate transformation. 

\textbf{Image Registration}: In \cite{SUN.2018}, a multi-stage approach for grid map registration is developed: At first, corner points are extracted from the OGM. According to those points, an initial optimal transformation matrix is calculated based on the isomorphism scheme of a triangle. This transformation is optimized by iterating the isomorphism scheme with more corner points and finding a maximum common subgraph. A similar approach is presented in~\cite{TANG.2021}, except that the feature point extraction and matching are performed using an image stitching CNN architecture in combination with a matching topology graph. These methods are suitable for the alignment but are not able to compensate for errors in the input OGMs.

\section{METHODOLOGY}

\begin{figure*}[!ht]
    \center
    \includegraphics[width=0.8\linewidth]{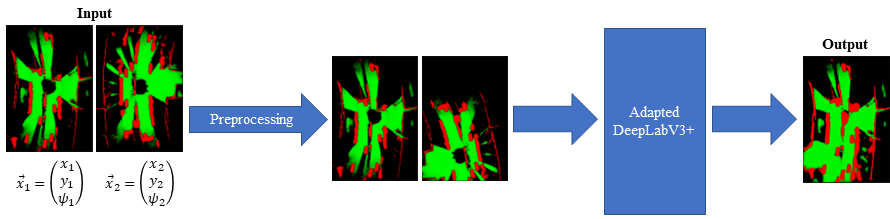}
    \caption{\label{flowchart}Two evidential OGMs perceived by different AVs are transformed into a common coordinate system based on the estimated vehicles poses and fed into the adapted DeepLabV3+, which predicts a fused OGM as output. During training, augmentation methods are applied.}
\end{figure*}

The proposed deep neural network processes two evidential OGMs and corresponding (uncertain) poses as input data and predicts the fused OGM from the perspective of the AV that perceived the first input OGM. Figure~\ref{flowchart} shows the data processing pipline during training and prediction time. In the following, the network architecture, synthetic training data generation, and the preprocessing steps comprising prealignment and augmentation are described.

\subsection{Network Architecture}

The architecture used in this work is based on DeepLabV3+~\cite{CHEN.2018}, which is a popular CNN architecture that was initially developed for image segmentation. The network consists of three main parts: The network backbone, the encoder, and the decoder. The encoder features atrous convolutions with different rates, point-wise convolutions and pooling operations. The decoder mainly features convolution and upsampling layers. 

In this work, ResNet-50~\cite{HE.2016} is used as the network's backbone and the last activation function of the network is changed to ReLU instead of softmax. Thus, for each cell $i$ the model is able to predict evidence for the cell being free $e_{i,F}$ and the cell being occupied $e_{i,O}$. These evidences $e_{i,A} \geq 0$ with $A \in \Theta$ can be transformed into belief masses $m_{i,A} \in [0,1]$ and an uncertainty mass $u_i \in [0,1]$:

\begin{eqnarray}
    \alpha_{i,A} & = & e_{i,A} + 1 \label{eq:evidence_to_alpha} \\
    m_{i,A} & = & e_{i,A} / S \\
    u_i & = & K / S_i
\end{eqnarray}

with the number of classes $K=2$ and the Dirichlet strength $S = \sum_{A \in \Theta} \alpha_{i,A}$.

The loss function used for this architecture is based on the loss function introduced in~\cite{VANKEMPEN.2021} with the following adaptions: (1) The Kullback-Leibler divergence term is neglected as it did not have a positive impact on the training progress and (2) a weighting factor $o_w \geq 1$ is introduced to account for the underrepresentation of occupied cells compared to free cells in the OGMs.

Each cell $i$ in the evidential OGM contains two belief masses, one for the cell being free $m_{i,F}$ and one for the cell being occupied $m_{i,O}$. In the following equation, $y_i$ is the true belief mass in the label while $\hat{p}_i$ is a parameter of a Dirichlet probability density function and can be computed from the evidences predicted by the neural network as $\hat{p}_{i,A} = \hat{\alpha}_{i,A} / S_i$ using Equation~\ref{eq:evidence_to_alpha}.

\begin{equation}
    L_i(w) = o_w \cdot \mathcal{L}_i(w), \;\;\;\; o_w = \left\{\begin{array}{ll} o_w, & y_{i,O} > 0.5 \\  1, & y_{i,O} \le 0.5\end{array}\right.
\end{equation}
with
\begin{equation}
\begin{split}
    \mathcal{L}_i(w)& = (y_{i,F} - \hat{p}_{i,F})^2 + \frac{\hat{p}_{i,F} (1 - \hat{p}_{i,F})}{S_i + 1} \\
                    & \quad + (y_{i,O} - \hat{p}_{i,O})^2 + \frac{\hat{p}_{i,O} (1 - \hat{p}_{i,O})}{S_i + 1}.
\end{split}
\end{equation}

\subsection{Training Data Generation}\label{trainingdata}

\begin{figure}[!ht]
    \center
    \includegraphics[width=0.9\linewidth]{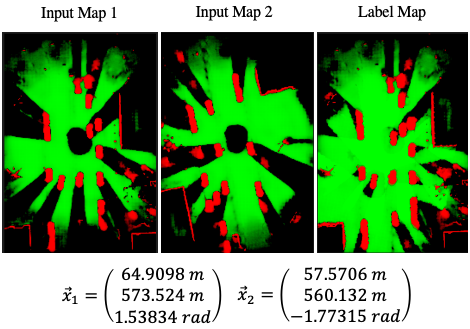}
    \caption{\label{inputlabelpair}A representative input-label pair consisting of two input maps with corresponding poses and the fused map from the perspective of the first input map.}
\end{figure}

Creating training data for the task at hand using real-world data would involve an immense labeling effort and is not easily scalable. Hence we decided to use synthetic training data, which has already shown good generalization capabilities for the task of occupancy grid mapping in our previous work~\cite{VANKEMPEN.2021}. Using an advanced simulation framework~\cite{vonNeumannCosel.2009} that provides a lidar plugin based on ray tracing and physically-based rendering, it becomes possible to generate as much training data as required virtually for free. In the simulation, we modeled one of our research vehicles that is equipped with a Velodyne VLP32C lidar sensor on its roof. We used two instances of this model in our simulations to generate lidar point clouds that are processed by the deep inverse sensor model presented in~\cite{VANKEMPEN.2021} to generate OGMs. If the two vehicles are close enough to each other, i.e. the OGMs are overlapping, the maps and the corresponding exact poses are stored as shown in Figure~\ref{inputlabelpair}. They make up the input data for the training. The label data consists of the correct fusion of both input OGMs. The fusion is performed twice, once from each of the two perspectives, and consists of several steps. Starting from the precisely known poses, both OGMs are transformed into a common coordinate system. After this coordinate transformation, the stored belief masses are fused cellwise using Dempster’s Rule of Combination (cf. Equation \ref{combination}). Afterwards, both fused OGMs are stored as labels. Since the fusions have been performed from both perspectives, the generated data can be split up into two input-label pairs.

\subsection{Prealignment}

Prealignment is the first step of the preprocessing. Based on the two input poses, a coordinate transformation is conducted that prealigns the second input map to the first one. Depending on the pose uncertainties, this alignment can be rather rough. As the poses are perfectly accurate in the simulation, artificial uncertainty is added to the data. The uncertainty is drawn from a Gaussian normal distributions, which is calculated separately for the $x$ and $y$ coordinates and for the orientation $\psi$. The Gaussian normal distributions have a zero mean and the $98\%$ confidence interval for the translational and rotational uncertainties are defined by $r$ and $\alpha$ respectively.

\subsection{Augmentation}

During the training, the input-label-pairs are augmented to increase the diversity of the training data and thus improve the generalization of the trained model. Random horizontal and vertical flips as well as random rotation in the interval from $-20$ to $20$ degrees are applied. After augmentation, both input OGMs are concatenated along the channel axis, so that each cell in the resulting map has four channels with free and occupied belief masses from both input OGMs. This concatenated tensor is used as input for the neural network.

\section{EXPERIMENTAL SETUP AND RESEARCH QUESTIONS}

As introduced in Section~\ref{trainingdata}, our deep learning-based registration and fusion model is trained using synthetic data, which is generated in a simulation environment. The simulated scenarios take place in an urban environment, where 9 templates are extracted that only cover small road sections of about 400 meters to make sure that both vehicles meet each other. The templates comprise different types of intersections as well as straight road segments. More than 600 scenarios were generated based on those 9 templates, each containing individual, random variations. Those variations include e.g. pulk traffic, parked cars, and moving pedestrians. The pulk traffic consists of a large variety of cars, trucks and motorcycles. A total of 10,000 input-label pairs were generated this way. That data was split into training, validation, and test dataset with a ratio of 80:10:10.

Four models with different configurations were trained, whereby each configuration differed in the amount of uncertainty that was applied to the input data. All models reached a minimum validation loss after approximately $200$ epochs with an initial learning rate of $0.01$, a learning rate decay factor of $0.5$ and a plateau patience of $10$ epochs before the decay factor is applied. The batch size was set to a maximum value of 80 and the occupation factor $o_w$ was set to an empirically determined value of $3.0$.

\begin{figure}[!ht]
\centering
    \renewcommand{\arraystretch}{1.5}
    \begin{tabular}{|l|c|c|c|c|}
    \hline
    Configuration & A & B & C & D \\
    \hline
    $r$ [m] & 0.0 & 1.0 & 2.5 & 5.0 \\
    \hline
    $\alpha$ [deg] & 0.0 & 10.0 & 15.0 & 20.0 \\
    \hline

    \end{tabular}
    \caption{\label{trainingconfiguration}Four different training configurations are used to train four different models. $r$ denotes the 98\% confidence interval for the translational uncertainty and $\alpha$ for the rotational uncertainty of the AVs' poses.}
\end{figure}

The architecture is trained using the Adam optimizer. The input OGMs have a length of $81.92$ meters and a width of $56.32$ meters. Their cells have a side length of $32$ centimeters, thus they have a size of $256$x$176$ pixels. The origin of the sensor is located in the middle of the OGMs.

In the following, we want to answer the research questions: 

\begin{itemize}
    \item How can a methodology for combined registration and fusion of evidential OGMs be designed?
    \item How well is our deep learning-based approach able to compensate for spatial misalignments of the input OGMs?
    \item How does its performance compare to a rule-based baseline approach for different degrees of misalignment in the input OGMs?
    \item How does the model trained on synthetic data perform when presented with real-world data?
\end{itemize}

\section{RESULTS AND DISCUSSION}

The four trained deep learning-based fusion models are evaluated qualitatively and quantitatively on the synthetic test dataset. The different trained models are compared to a baseline two-stage fusion approach that consists of coordinate transformation in the first stage and belief mass fusion using Dempster's Rule of Combination (cf. Equation~\ref{combination}). Eventually, one of the models trained on synthetic data is evaluated qualitatively on real-world data to analyze its generalization capabilities after a domain shift.

\subsection{Evaluation on Synthetic Data}

\begin{figure}[!ht]
    \center
    \includegraphics[width=0.9\linewidth]{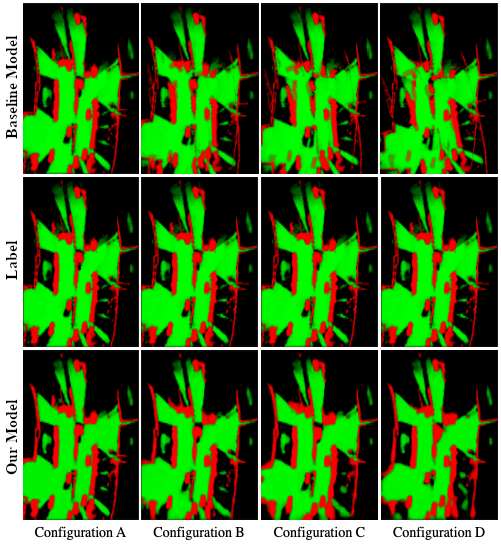}
    \caption{\label{qualitative}Fusion results our models (bottom) compared to the results of the baseline model (top) for the different configurations as defined in Figure~\ref{trainingconfiguration}. The applied pose uncertainties are set to 80~\% of the defined 98~\% confidence interval.}
\end{figure}

Representative prediction results are depicted in Figure~\ref{qualitative}. The prediction results of the deep learning-based models are contrasted with the label and the naive fusion. For the purpose of this visualization, 80~\% of the scenario-specific 98~\% confidence intervals are applied to the input poses. As can be seen, both the naive fusion and the deep learning-based model perform very well for configuration A. Starting with configuration B, the naive fusion becomes worse with larger amounts of uncertainty. Street limits and parked as well as moving cars are heavily misaligned. The deep learning-based models, on the other hand, produce much more reliable and consistent results. Occupied areas are somewhat larger and predictions get less accurate with growing uncertainties, but most of the important areas are still covered and localized correctly, even if high uncertainties are applied.

\begin{figure}[!ht]
    \center
    \includegraphics[width=\linewidth]{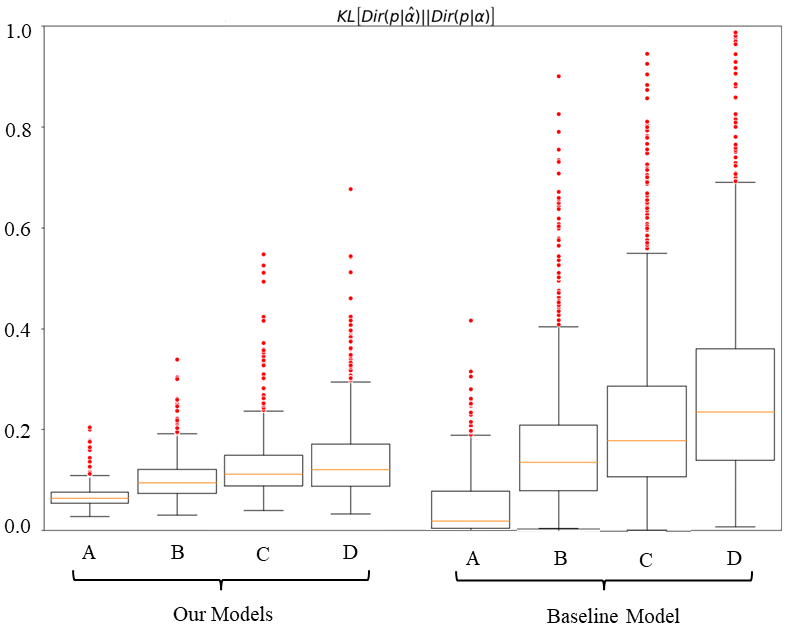}
    \caption{\label{kld}KLD score distribution over the test dataset for our models and the corresponding results using the baseline model according to the training configuration depicted in Figure~\ref{trainingconfiguration}.}
\end{figure}

Figure~\ref{kld} depicts boxplots of the Kullback-Leibler divergence (KLD) distributions of both results from the deep learning-based models and from the naive fusion for the different training configurations over the test dataset. Expectantly, the KLD distributions become worse, the more uncertainty is applied. But it is striking that the decline of fusion quality happens much faster and stronger for the naive fusion. While the initial deep learning-based model is, on average, worse than the naive fusion, all the following models are much better. This diagram not only shows that the average prediction quality of both approaches is drifting apart but also that the prediction quality of the deep learning-based model varies much less, and is thus more reliable.

To be able to calculate classification scores, the predicted belief masses need to be transformed into a categorical representation. It is assumed that each cell has one of the two classes "free" and "occupied". At first, the predicted masses are transformed to classical probabilities: $p_{F}$ and $p_{O}$ using Equation~\ref{eq:pignistic}. Then, the classification is performed: A cell is denoted free if $p_{O} > 0.5$, otherwise occupied.

\begin{figure}[!ht]
    \centering
	\begin{tikzpicture}
	\begin{axis}[
 ybar,
 xtick={1,2,3,4},
 xticklabels={A, B, C, D},
 ymin=0.92, ymax=1.0, 
 bar width=3,
 legend entries={Precision (Our), Precision (Baseline), Recall (Our), Recall (Baseline), Dice (Our), Dice (Baseline)},
 legend style={at={(0.5,-0.2)},anchor=north},
 legend cell align={left},
 legend columns=2,
 width=\linewidth,
 height=5cm,
 ]
	\addplot+[blue!100!white] coordinates
	{(1.0, 0.991) (2.0, 0.988) (3.0, 0.986) (4.0, 0.982) };
	\addplot+[blue!40!white] coordinates 
	{(1.0, 0.993) (2.0, 0.979) (3.0, 0.97) (4.0, 0.964) };
	\addplot+[green!100!white!60!black] coordinates 
	{(1.0, 0.97) (2.0, 0.961) (3.0, 0.957) (4.0, 0.955) };
	\addplot+[green!40!white!80!black] coordinates 
	{(1.0, 0.991) (2.0, 0.975) (3.0, 0.967) (4.0, 0.961) };
	\addplot+[red!100!white!60!black] coordinates 
	{(1.0, 0.98) (2.0, 0.974) (3.0, 0.971) (4.0, 0.968) };
	\addplot+[red!40!white!80!black] coordinates 
	{(1.0, 0.991) (2.0, 0.977) (3.0, 0.968) (4.0, 0.962) };

    \end{axis}
	\end{tikzpicture}
	\caption{\label{fig_free}Classification scores for free cells predicted by our model compared to the baseline model.}
\end{figure}
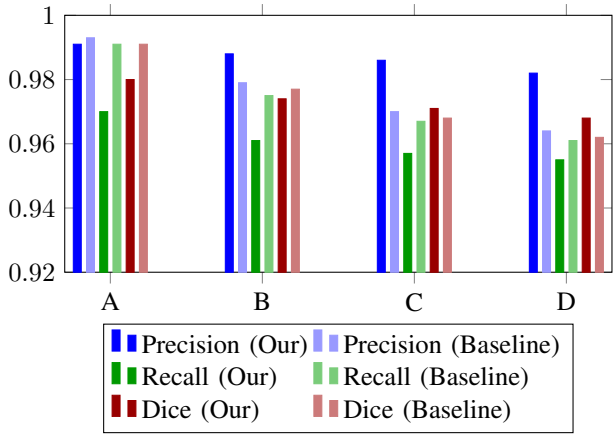

The Figures~\ref{fig_occupied} and~\ref{fig_free} show the evolution of the precision, recall, and dice scores for the deep learning-based models and the corresponding naive fusions. The first diagram depicts the scores for the occupied cells and the second one for the free cells. As expected, all scores gradually decrease with increasing uncertainty. The slopes of the curves belonging to the deep learning-based models, however, are all much flatter. This suggests that uncertainties have a greater impact on the naive fusion. Without any uncertainties, all scores are better for the naive fusion. But as uncertainties increase, the deep learning-based models eventually become dominant in most categories. Merely the precision for occupied cells and the recall for free cells is worse, which is the consequence of factor $o_w$ that was set to 3 and emphasizes the model to predict a higher belief mass for the occupied compared to the free state.

\begin{figure}[!ht]
    \centering
	\begin{tikzpicture}
	\begin{axis}[
 ybar,
 xtick={1,2,3,4},
 xticklabels={A, B, C, D},
 ymin=0.92, ymax=1.0, 
 bar width=3,
 legend entries={Precision (Our), Precision (Baseline), Recall (Our), Recall (Baseline), Dice (Our), Dice (Baseline)},
 legend style={at={(0.5,-0.2)},anchor=north},
 legend cell align={left},
 legend columns=2,
 width=\linewidth,
 height=5cm,
 ]
    \addplot+[blue!100!white] coordinates
	{(1.0, 0.949) (2.0, 0.943) (3.0, 0.929) (4.0, 0.927) };
	\addplot+[blue!40!white] coordinates 
	{(1.0, 0.984) (2.0, 0.961) (3.0, 0.949) (4.0, 0.94) };
	\addplot+[green!100!white!60!black] coordinates 
	{(1.0, 0.984) (2.0, 0.98) (3.0, 0.977) (4.0, 0.971) };
	\addplot+[green!40!white!80!black] coordinates 
	{(1.0, 0.989) (2.0, 0.969) (3.0, 0.957) (4.0, 0.948) };
	\addplot+[red!100!white!60!black] coordinates 
	{(1.0, 0.966) (2.0, 0.956) (3.0, 0.952) (4.0, 0.948) };
	\addplot+[red!40!white!80!black] coordinates 
	{(1.0, 0.986) (2.0, 0.965) (3.0, 0.953) (4.0, 0.944) };

    \end{axis}
	\end{tikzpicture}
	\caption{\label{fig_occupied}Classification scores for occupied cells predicted by our model compared to the baseline model.}
\end{figure}
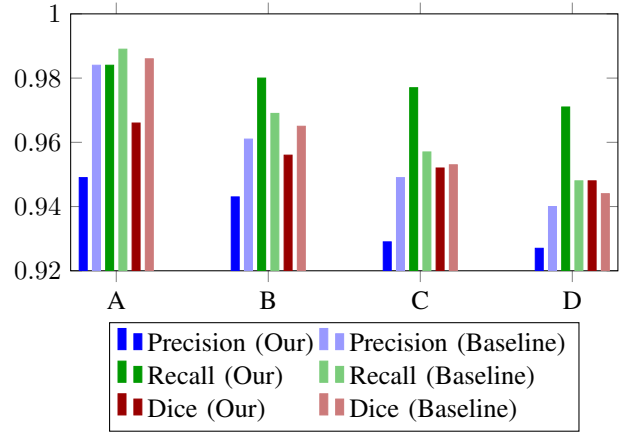

\subsection{Evaluation on Real-World Data}

\begin{figure}[!ht]
    \center
    \includegraphics[width=0.9\linewidth]{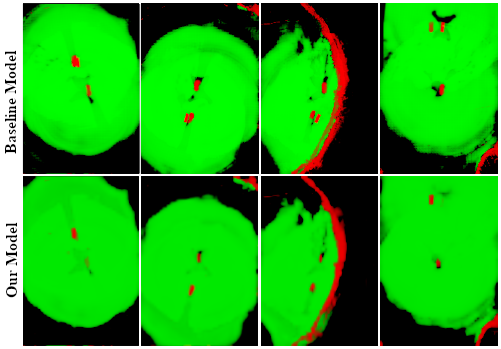}
    \caption{\label{img:evaluation-real}The bottom row shows the results of our proposed model when presented with OGMs recorded with two of our research vehicles on the test track compared to results of the baseline approach in the top row.}
\end{figure}

Model D (cf. Figure~\ref{trainingconfiguration}), which was also used for the evaluation on synthetic data before, was also tested with real-world data recorded with two of our research vehicles. Both vehicles have a 32-layer lidar sensor on their roofs at approximately the same position. The deep inverse sensor model presented in \cite{VANKEMPEN.2021} is used to compute OGMs from the lidar data and the vehicles use GNSS to estimate their poses. The OGMs and poses recorded in both vehicles were manually synchronized in time and fed into the proposed DNN. Figure \ref{img:evaluation-real} shows the output of the DNN compared to the results of the baseline approach.

\section{CONCLUSION AND OUTLOOK}

We presented a methodology that uses a CNN to spatially align and fuse two evidential OGMs perceived by different AVs. The approach is able to compensate for errors both in the vehicle poses as well as in the OGMs. The presented model is trained with synthetic data and outperforms a baseline approach as soon as the poses of the AVs are affected by errors. The experiments with synthetic data have shown that the deep learning-based model shows good results even for spatial misalignments of up to 5 meters and 20 degrees between both OGMs. A qualitative evaluation on real-world data demonstrates the generalization capabilities of the model to real-world data.

However, the deep learning-based model tends to overestimate occupied regions leading to a lower recall for the estimation of free cells and a lower precision for the estimation of occupied cells. First experiments with OGMs of higher resolution have shown potential to overcome this problem and will be analyzed in future work.

\section*{Acknowledgment}

This research is accomplished within the project ”AUTOtech.\textit{agil}” (FKZ 01IS22088A). We acknowledge the financial support for the project by the Federal Ministry of Education and Research of Germany (BMBF).

\bibliographystyle{IEEEtran}
\bibliography{literature}

\begin{thebibliography}{10}
\providecommand{\url}[1]{#1}
\csname url@samestyle\endcsname
\providecommand{\newblock}{\relax}
\providecommand{\bibinfo}[2]{#2}
\providecommand{\BIBentrySTDinterwordspacing}{\spaceskip=0pt\relax}
\providecommand{\BIBentryALTinterwordstretchfactor}{4}
\providecommand{\BIBentryALTinterwordspacing}{\spaceskip=\fontdimen2\font plus
\BIBentryALTinterwordstretchfactor\fontdimen3\font minus
  \fontdimen4\font\relax}
\providecommand{\BIBforeignlanguage}[2]{{%
\expandafter\ifx\csname l@#1\endcsname\relax
\typeout{** WARNING: IEEEtran.bst: No hyphenation pattern has been}%
\typeout{** loaded for the language `#1'. Using the pattern for}%
\typeout{** the default language instead.}%
\else
\language=\csname l@#1\endcsname
\fi
#2}}
\providecommand{\BIBdecl}{\relax}
\BIBdecl

\bibitem{Lampe.2020}
B.~Lampe, R.~{van Kempen} \emph{et~al.}, ``Reducing uncertainty by fusing
  dynamic occupancy grid maps in a cloud-based collective environment model,''
  in \emph{2020 IEEE Intelligent Vehicles Symposium (IV)}, 2020, pp. 837--843.

\bibitem{autotech}
``The autotech.agil project,''
  \url{https://www.ika.rwth-aachen.de/autotechagil}, accessed: 2022-12-29.

\bibitem{YUE.2016}
Y.~Yue, D.~Wang \emph{et~al.}, ``A hybrid probabilistic and point set
  registration approach for fusion of 3d occupancy grid maps,'' \emph{2016 IEEE
  International Conference on Systems, Man, and Cybernetic}, pp. 28--34,
  October 9-12, 2016.

\bibitem{MENG.2022}
X.~Meng, D.~Duan, and T.~Feng, ``Multi-vehicle multi-sensor occupancy grid map
  fusion in vehicular networks,'' \emph{IET Communications 16}, pp. 67--74,
  2022.

\bibitem{LI.2018}
Y.~Li, D.~Duan \emph{et~al.}, ``Occupancy grid map formation and fusion in
  cooperative autonomous vehicle sensing,'' \emph{2018 IEEE International
  Conference on Communication Systems (ICCS)}, 2018.

\bibitem{MARKIEWICZ.2018}
P.~Markiewicz, K.~Kogut \emph{et~al.}, ``Occupancy grid fusion prototyping
  using automotive virtual validation environment,'' \emph{ICCMA 2018:
  Proceedings of the 6th International Conference on Control, Mechatronics and
  Automation}, pp. 81--85, 2018.

\bibitem{SUN.2018}
Y.~Sun, R.~Sun \emph{et~al.}, ``A grid map fusion algorithm based on maximum
  common subgraph,'' \emph{Proceedings of the 2018 13th World Congress on
  Intelligent Control and Automation}, July 4-8, 2018.

\bibitem{TANG.2021}
Q.~Tang, K.~Zhang \emph{et~al.}, ``Map fusion method based on image stitching
  for multi-robot slam,'' \emph{Advances in Swarm Intelligence: 12th
  International Conference, ICSI 2021}, vol. 810, pp. 146--154, 2021.

\bibitem{Elfes.1989}
A.~Elfes, ``Using occupancy grids for mobile robot perception and navigation,''
  \emph{Computer}, vol.~22, no.~6, pp. 46--57, Jun. 1989.

\bibitem{Thrun.2005}
S.~Thrun, W.~Burgard, and D.~Fox, \emph{Probabilistic robotics}, ser.
  Intelligent robotics and autonomous agents.\hskip 1em plus 0.5em minus
  0.4em\relax Cambridge, Mass.: {MIT Press}, 2005.

\bibitem{Dempster.1968}
A.~P. Dempster, ``A generalization of bayesian inference,'' \emph{Journal of
  the Royal Statistical Society: Series B (Methodological)}, vol.~30, no.~2,
  pp. 205--232, Jul. 1968.

\bibitem{Shafer.1976}
G.~Shafer, \emph{A mathematical theory of evidence}, ser. Limited paperback
  editions.\hskip 1em plus 0.5em minus 0.4em\relax Princeton, NJ: {Princeton
  Univ. Press}, 1976, vol.~42.

\bibitem{NUSS.2013}
D.~Nuss, B.~Wilking \emph{et~al.}, ``Decision-free true positive estimation
  with grid maps for multi-object tracking,'' \emph{Proceedings of the 16th
  International IEEE Annual Conference on Intelligent Transportation Systems
  (ITSC 2013)}, pp. 28--34, October 6-9, 2013.

\bibitem{VANKEMPEN.2021}
R.~{Van Kempen}, B.~Lampe \emph{et~al.}, ``A simulation-based end-to-end
  learning framework for evidential occupancy grid mapping,'' \emph{2021 IEEE
  Intelligent Vehicles Symposium (IV)}, 2021.

\bibitem{Lang.2019b}
A.~H. Lang, S.~Vora \emph{et~al.}, ``Pointpillars: Fast encoders for object
  detection from point clouds,'' in \emph{2019 IEEE/CVF Conference on Computer
  Vision and Pattern Recognition (CVPR)}, 2019, pp. 12\,689--12\,697.

\bibitem{CHEN.2018}
L.~Chen, Y.~Zhu \emph{et~al.}, ``Encoder-decoder with atrous separable
  convolution for semantic image segmentation,'' \emph{Proceedings of the
  European conference on computer vision (ECCV)}, pp. 801--818), 2018.

\bibitem{HE.2016}
K.~He, X.~Zhang \emph{et~al.}, ``Deep residual learning for image
  recognition,'' \emph{Proceedings of the IEEE conference on computer vision
  and pattern recognition}, pp. 770--778, 2016.

\bibitem{vonNeumannCosel.2009}
K.~{von Neumann-Cosel}, M.~Dupius, and C.~Weiss, ``Virtual test drive -
  provision of a consistent tool-set for [d,h,s,v]-in-the-loop,'' in
  \emph{Proceedings of the driving simulation conference Monaco}, 2009.

\end{thebibliography}

\vspace{12pt}

\end{document}